\documentclass[12pt]{article}
\usepackage{scicite}
\usepackage{times}
\usepackage{epsfig,color,graphics}
\usepackage{siunitx}
\usepackage{amsmath, mathrsfs}
\usepackage{comment}
\usepackage{upgreek}
\usepackage[normalem]{ulem}
\usepackage{float}
\usepackage[latin1]{inputenc}

\usepackage{xcolor}

\newenvironment{sciabstract}{%
	\begin{quote} \bf}
	{\end{quote}}

\title{\textbf{A self-organizing robotic aggregate using solid and liquid-like collective states}}

\author{Baudouin Saintyves,$^{1\ast}$ Matthew Spenko,$^{2}$ Heinrich M. Jaeger$^{1,3}$ \\
	\\
	\normalsize{$^{1}$James Franck Institute, University of Chicago, Chicago, IL 60637, USA}\\
	\normalsize{$^{2}$Mechanical, Materials, and Aerospace Engineering},\\ \normalsize{Illinois Institute of Technology, Chicago, IL 60614, USA}\\
	\normalsize{$^{3}$Department of Physics, University of Chicago, Chicago, IL 60637, USA}\\
	\\
	\normalsize{$^\ast$To whom correspondence should be addressed; E-mail: saintyves@uchicago.edu.}
}
\date{}
\begin{document}
	\baselineskip24pt
	\maketitle
	
	\begin{sciabstract}
		
		Designing robotic systems that can change their physical form factor as well as their compliance to adapt to environmental constraints remains a major conceptual and technical challenge. To address this, we introduce the Granulobot, a modular system that blurs the distinction between soft, modular, and swarm robotics. The system consists of gear-like units that each contain a single actuator such that units can self-assemble into larger, granular aggregates using magnetic coupling. These aggregates can reconfigure dynamically and also split into subsystems that might later recombine. 
		Aggregates can self-organize into collective states with solid- and liquid-like properties, thus displaying widely differing compliance. 
		These states can be perturbed locally via actuators or externally via mechanical feedback from the environment to produce adaptive shape shifting in a decentralized manner. This in turn can generate locomotion strategies adapted to different conditions. Aggregates can move over obstacles without using external sensors or coordinate to maintain a steady gait over different surfaces without electronic communication among units. The modular design highlights a physical, morphological form of control that advances the development of resilient robotic systems with the ability to morph and adapt to different functions and conditions.
	\end{sciabstract}
	
	\section*{Introduction}
	
	The development of autonomous and efficient agents that can adapt to a variety of environments and perform different functions by reconfiguring their body is one of the frontiers in the field of robotics. There is an increasing need for systems that offer multi-functional, self-assembling, highly compliant capabilities in concert with resilience and robustness \cite{Gao2017}. To this end, while traditional robotic design tends to distinguish and separate components dealing with sensing, actuation, computation, and communication, an alternative strategy distributes components that each integrate these functions \cite{McEvoy2015}. For instance, modular design approaches replace a single mechanical body with an aggregate of multi-functional subunits that can assemble and couple together \cite{Romanishin2013,Romanishin2015,Piranda2018,Liang2020}. However, this coupling often leads to rigid connections between docked units, which then implies that changes of the overall aggregate shape require an iterative reconfiguration process whereby units need to disconnect, move, and then reconnect at another location. This is incompatible with a variable mechanical compliance of the overall structure and limits adaptability to changing tasks or environments. 
	
	Soft robotics may overcome some of these limitations \cite{Brown2010,Tolley2014,Paik2011,Felton2014,Howell2013,Polygerinos2015,Proietti2021,Gazzola2018,Li2017}. Still, while robotic systems with a single soft body are able to achieve highly variable and adaptive behaviors \cite{Roche2017}, the inherent need to model materials in the large deformation limit makes the design difficult. 
	Furthermore, inspired by flocking observed in nature \cite{Vicsek2012,Attanasi2014}, multi-agent swarm approaches have been successful in generating autonomous, as well as resilient robotic systems by using a large number of individual robotic units where large-scale collective behaviors emerge through local rules between neighboring units \cite{Theraulaz2020,Rubinstein2014,Caprari2001,Giomi2013,BenZion2022}. Such non-centralized control is able to absorb local perturbations in the inherent noise of the collective organization, and the failure of a few in a swarm of similar units does not necessarily affect the capabilities at the collective scale \cite{Rubinstein2014,Werfel2014}. However, robotic swarms have so far been limited to low density, fluid-like systems with very little overall rigidity of the system as a whole.    
	
	\begin{figure}
		\centering
		\includegraphics[width=\textwidth]{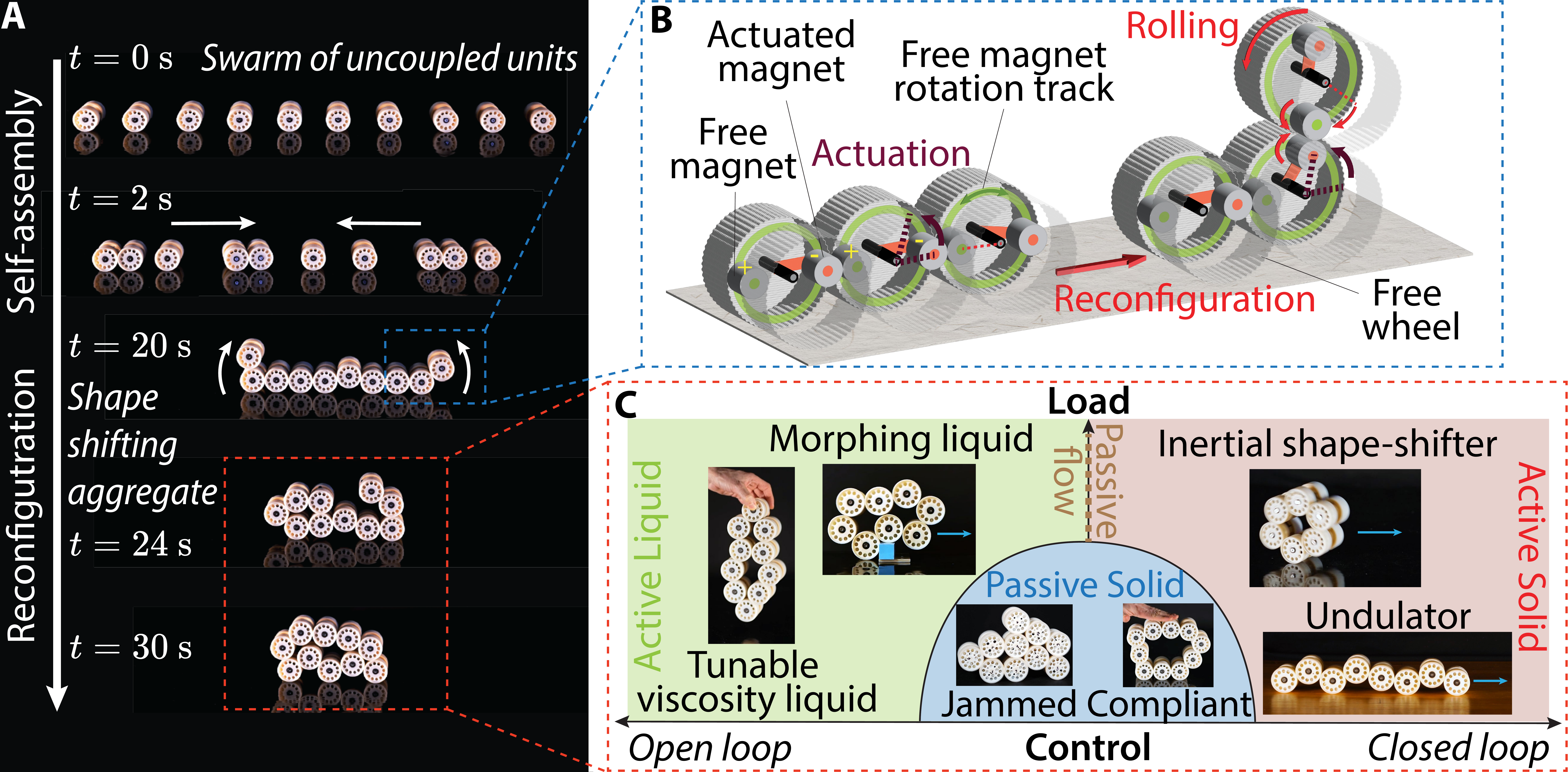}
		\caption{\textbf{Granulobots: A modular platform bridging soft robotics and active granular materials}. See Movie S1. (A) Time-lapse of a swarm of Granulobot units self-assembling into an aggregate and then reconfiguring. (B) Coupling and actuation mechanism inside the Granulobot units.  Each two-wheeled unit uses magnetic coupling to connect to its neighbors. Inside a given unit, a single, rotary degree of freedom can be activated to generate torque, which enables this unit to roll along the frictional contact with its neighbor. (C) Granulobot aggregate behaviors, ranging from active liquid to active solid, depending on the type of control.  Controlling in open-loop enables behaviors resembling an active liquid with tunable viscosity and the ability to move and morph. In closed loop  individual units employ local, onboard feedback and the aggregate can exhibit a range of solid-like behaviors with locomotion gaits that include collective rolling and synchronized undulation. For small applied mechanical loads and sufficiently small bias voltages to the units,  friction maintains the aggregate in a passive state that can resemble jammed granular matter or an elastic solid, depending on the units' spatial arrangement.}
		\label{fig_1}
	\end{figure}
	
	The system described in this paper, Granulobots, combines several desirable features of modular, soft, and multi-agent swarm robotics. The Granulobot design is inspired by granular matter's highly strain-adaptive behavior and ability to transform between rigid and fluid-like states via a jamming transition \cite{Jaeger2015}. The remarkable features of granular matter arise from collective effects in assemblies made of simple building blocks and depend on inter-particles contact properties. Expanding beyond granular, jamming-based granular actuator systems \cite{Brown2010,Steltz2010,Cheng2012,Cianchetti2014,Tramsen2020,Akta2021,Wang2021} and gear-based metamaterials \cite{Meeussen2016,Shaw2018,Fang2022}, replacing passive grains with motorized particles adds new options for locomotion \cite{Savoie2019,Batra2019,Agrawal2020,Li2021,Karimi2022}. 
	In granular systems, overall malleable behavior can be achieved even with individually rigid sub-units by designing loose and detachable coupling \cite{Batra2019}. With the Granulobots, we introduce a novel coupling design that enables continuous deformations and dynamical coupling in aggregates of self-assembling motorized units. In dense assemblies of multiple Granulobots, this allows for both solid- and liquid-like collective behaviors under a gravity field, which then can be exploited for a range of locomotion strategies. The result is a new form of active granular material that can, at the same time, form the building blocks of a soft robot.
	
	A swarm of Granulobot units, each with a single actuated degree of freedom that enables it to roll like a wheel on the ground, can self-assemble into larger granular aggregates (see Fig.\ \ref{fig_1}A, Movie S1). 
	Units can then shift their positions simultaneously to generate arbitrary overall shapes. An individual Granulobot unit consists of a cylindrical body with an embedded control circuit and two permanent magnets that can rotate about the cylinder's axis. One magnet rotates freely, while the other one is actuated with a motor. This design allows individual units to move autonomously, mechanically connect with neighbors via magnets, form a continuously deformable aggregate, and apply torque to a neighbor (Fig.\ \ref{fig_1}B). The aggregate's ability to reorganize its global torque state dynamically allows for collective shape-shifting into the vertical plane and against gravity.
	
	Two modes of controlling the behavior of individual units provide access to a rich set of aggregate behaviors, with responses ranging from rigid and solid to viscous and liquid-like (Fig.\ \ref{fig_1}C). 
	Coupled with feedback control, an onboard encoder sensor helps each Granulobot unit maintain the angular position of its actuated rotor, which leads to a solid-like, i.e., stiff elastic, response of the aggregate. Feedback control can also be used to generate spontaneous, self-organized oscillations. This makes it possible to implement locomotion strategies based on collective rolling or rhythmic undulation \cite{Ijspeert2007,Zhou2021,Brandenbourger2022}. In contrast, when employing open loop control, the angular position of the rotors is not maintained, allowing for irreversible deformation of the aggregate in response to mechanical load. This can be exploited to create a wide range of fluid-like behaviors of the aggregate, including locomotion, which cause the robot to act like a soft system that yields and conforms to external forces. Importantly, without applied power, a state with structural resistance can be maintained due to a static friction threshold. This is in contrast to other state-changing materials such as electro- and magnetorheological fluids that require large power consumption to maintain a rigid state, which limits their use in autonomous robotics \cite{Wen2003}. 
	
	\section*{Results}
	
	\paragraph*{\textbf{Granulobot Design Principle}\\}
	An individual Granulobot unit is itself an autonomous robot, cylindrical in shape with an internal actuator that controls a single rotary degree of freedom. It can magnetically couple to two other units to form more complex aggregates that can change their shape (Fig.\ \ref{fig_1}A). The actuator, a brushed DC motor, rotates a permanent magnet around a central axis as shown in  Fig.\ \ref{fig_1}B (actuated magnet). A second permanent magnet is attached to a passive, freely rotating shaft (free magnet) that can move at the same fixed radius around the cylinder's central axis. 
	When units come into close proximity, the magnetic attraction couples an actuated magnet with a free magnet, which allows the units to transmit torque between themselves. When the motor drives the actuated magnet to rotate (brown arrow in Fig.\ \ref{fig_1}B), the magnetic coupling forces the free magnet to follow by rotating around the robot central axis. Since Granulobots are designed with frictional, smooth gear-like surfaces, the produced torques can thus be transferred from the neighbor's free magnet to its robot body. This results in one unit rotating around the other (red arrow in Fig.\ \ref{fig_1}B).
	
	The actuated magnet also acts as an eccentric weight that, when rotated, shifts the unit's center of gravity and causes an isolated Granulobot to roll. This enables locomotion of individual units and makes it possible for them to autonomously assemble into an aggregate once they come within the range of the magnetic coupling force (Fig.\ \ref{fig_1}A). Docking is ensured by using opposite polarities for the magnets. Two possibilities arise: either both magnets have the same polarity in a given unit, in which case units with one polarity dock with units of the other polarity, or all units are identical with one polarity for the actuated magnet and the opposite polarity for the free magnet as shown in Fig.\ \ref{fig_1}B. All results presented here are valid for both cases. Since the free magnet is not attached to any other structure, it slides freely along a circular track. Although we limit the present study to a connectivity of two units, i.e. each unit is magnetically coupled to no more than two neighbors, in principle this design allows for the possibility of adding additional free magnets  to enable more complex aggregate configurations (see Fig. S1 for units that can couple to three neighbors). 
	
	Once aggregates are formed, the system is able to reconfigure and control its behaviors using decentralized algorithms enabled through robot-robot interactions. Each Granulobot contains a battery-powered circuit with Wi-Fi capabilities (see Materials and Methods) that can be used for real-time robot-robot communication. For the work discussed here, Wi-Fi communication with a central computer is only used to gather data for post-analysis and for sending initialization commands to all robots. All control strategies are autonomously and decentrally implemented by the units through their interactions.

	\begin{figure}
		\centering
		\includegraphics[width=\textwidth]{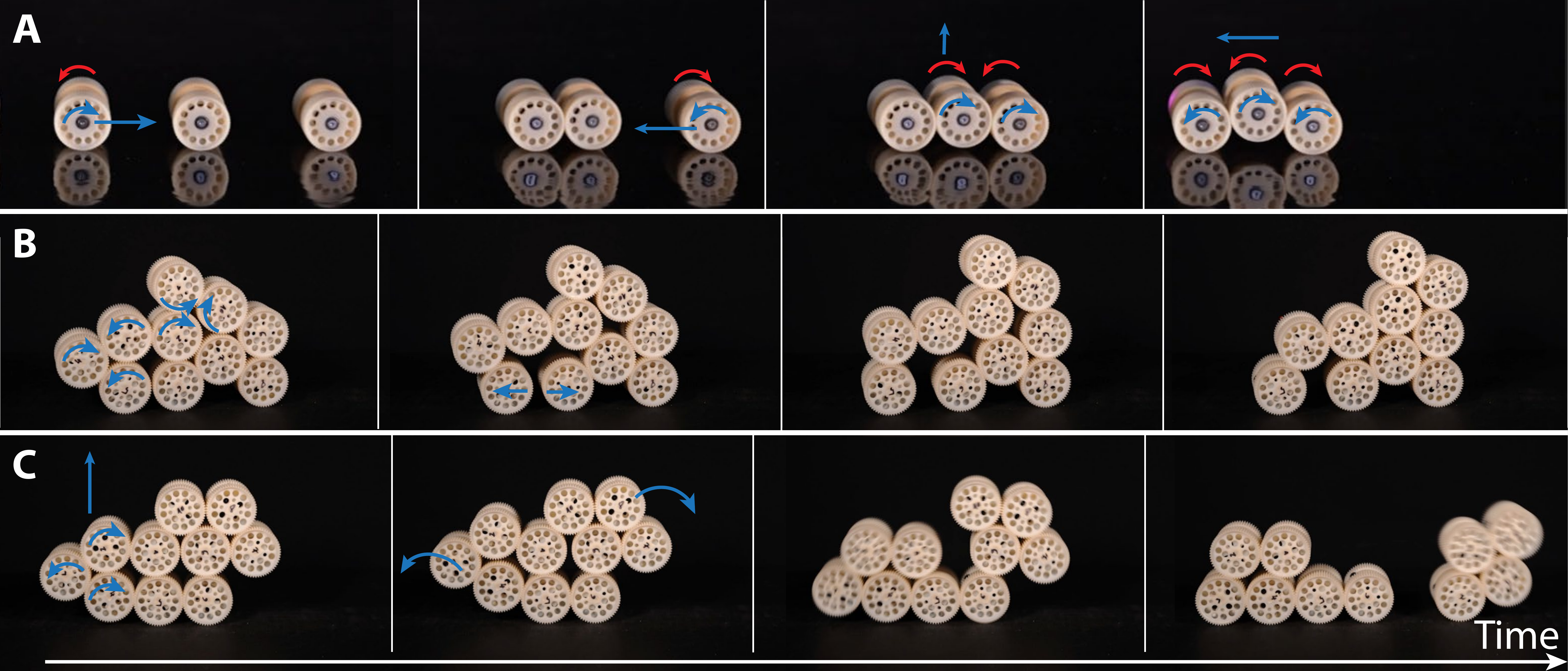}
		\caption{\textbf{Self-assembly, shape-shifting and dynamic reconfiguration.} Red arrows represent the actuated magnets' direction of rotation. Blue arrows represent Granulobots in the process of reconfiguration. (A) Individual Granulobot units can roll and attach magnetically into larger assemblies, which then can move using a subset of units as wheels. (B) Exerting torque onto their neighbors, individual units as well as groups of units can reposition themselves and thus rearrange the assembly's} shape. (C) By exerting torque larger than the magnetic binding between neighbors, units can split off and form autonomous robots on their own (See Movie S1).
		\label{fig_2}
	\end{figure}
	
	\paragraph*{\textbf{Control of Individual Units and Aggregate Formation / Disassembly}\\ } 
	
	In each individual Granulobot unit \emph{i}, torque balance with respect to its actuated rotor leads to the following relation among the rotor's angular speed $\dot{\theta_i}$; the torque $\Gamma_{\textrm{l},i}$ produced by an external load (e.g., from a neighboring unit or external perturbation); the voltage $U_i$ applied to the motor, which produces a torque $kU_i$; and the frictional torque from the motor's and unit's moving parts, which resists rotation and has magnitude $\Gamma_\textrm{f} > 0$ (taken to be the same for all units)
	
	\begin{equation}
		\eta_0\dot{\theta_i}=
		\left\{
		\begin{array}{ll}
			0 &    |\Gamma_{\textrm{l},i} +kU_i| \leq \Gamma_\textrm{f} \\
			\Gamma_{\textrm{l},i}+kU_i - \Gamma_\textrm{f}\mathrm{sgn}(\dot{\theta_i})    & |\Gamma_{\textrm{l},i} +kU_i|>  \Gamma_\textrm{f}\\
		\end{array} 
		\right. 
		\label{SteadyMotion1}
	\end{equation}
	where $\eta_0\dot{\theta_i}$ is the torque produced by the motor's back Electro-Motive Force (EMF) and $k > 0$ and $\eta_0 > 0 $ are parameters characterizing the motor's load response (see Materials and Methods for the derivation of Eq.\ \ref{SteadyMotion1} and the sign convention for the torques). 
	
	As long as $|\Gamma_{\textrm{l},i} +kU_i| <  \Gamma_\textrm{f}$, a Granulobot is effectively in a passive state and does not move. 
	For initially unconnected units in a swarm with $\Gamma_l=0$, a  voltage $U_i \geq \Gamma_\textrm{f}/k$ is required to start autonomous rolling, and enable units to connect with each other. Once the units have self-assembled, a change in the torque that a given unit exerts with its active rotor on its neighbor can lead the whole assembly to reorganize. As soon as an aggregate comprises three or more units, shape-shifting can enable collective locomotion. For example, in the right image of Fig.\ \ref{fig_2}A, the middle unit has lifted off the ground, which allows the aggregate to move (here to the left) by having all wheels that touch the ground rotate in the same direction. 
	
	In larger aggregates, applying torque can physically move units on top of others and deform the assembly into an arbitrary configuration within the vertical plane (Fig.\ \ref{fig_2}B, Movie S1). At the points where units contact each other, magnetic attraction provides a normal force strong enough to ensure no-slip gear-like coupling, while units that touch in the absence of magnetic attraction can slide as long as forces exceed the static friction arising from the units' smooth gear surface. These two frictional behaviors prevent units from seizing when three consecutive units  are in contact with each other. 
	
	During such reconfiguration, geometrical constraints that prevent a unit from rotating about its neighbor's central axis can still arise. If further rotation is nevertheless forced and the actuation torque applied becomes stronger than the magnetic coupling, the actuated magnet will separate from its counterpart in the neighboring unit. An ability for individual units to detach thus emerges from the collective properties of the assembly, without the use of any additional actuator. This allows the reorganization of neighbors in aggregates, and the creation of separate entities that can function as autonomous robots (Fig.\ \ref{fig_2}C, Movie S1).
	
	Choosing a functional form for the voltage $U_i$ in Eq. \ref{SteadyMotion1} makes it possible to generate different behaviors for the individual Granulobot units and therefore to design a rich set of collective mechanical behaviors for the aggregate. In the following equation, we use a general form for each unit $i$ that combines constant torque and position control by setting a voltage bias $u_i$ and monitoring the angular position $\theta_i$ from an encoder embedded in each unit's circuitry:
	
	\begin{equation}
		U_i= u_i-\left(\alpha~\frac{\eta_\textrm{0}}{k} \dot{\theta}_i +\frac{G}{k}\theta_i \right).
		\label{control}
	\end{equation}
	
	The parameters $G/k > 0$ and $\alpha \eta_\textrm{0}/k$, with  $\alpha \in [-1,\infty[$, determine how the drive voltage responds to changes in the rotor's movement, and can be thought of as the coefficients of a proportional-derivative (PD) feedback loop ($K_\textrm{p}$ and $K_\textrm{d}$, respectively \cite{Hughes1990}). Using such active control, a unit's electro-mechanical behavior for $|\Gamma_{\textrm{l},i} +kU_i|\geq  \Gamma_\textrm{f}$ in Eq. 1, i.e. beyond the passive state, is then described by
	
	\begin{equation}
		ku_i +\Gamma_{\textrm{l},i} - \Gamma_\textrm{f}\mathrm{sgn}(\dot{\theta_i})=(1+\alpha)\eta_\textrm{0} \dot{\theta}_i + G\theta_i.
		\label{SteadyState}
	\end{equation}
	
	Here $G$ tunes the Granulobot's ability to maintain its rotor at a targeted position under load, thus acting analogous to a torsion spring constant, while $(1+\alpha)\eta_\textrm{0}$ tunes the rotor's damping, analogous to viscous dissipation; $\alpha > 0$ increases the damping generated by the back emf $\eta_\textrm{0}\dot{\theta_i}$, while $\alpha = -1$ actively cancels it. $ku_i$ corresponds to a torque bias produced by the motor. 
	
	Within this framework, driving Granulobots in `open loop' in Fig.\ \ref{fig_1}C corresponds to setting parameters $\alpha = G = 0$, i.e. ignoring onboard sensor information about $\theta$ and $\Dot{\theta}$ and controlling the system only via the bias voltages $u_i$, while `closed loop' refers to $\alpha \geq -1$ and $G > 0$ so that the sensor data informs the response. In the next section, we describe how choices for these parameters control the aggregate behavior indicated in Fig.\ \ref{fig_1}C. In a subsequent section we then demonstrate how collective states can be leveraged for different aggregate locomotion control strategies that can be used in different environmental conditions.
	
	\begin{figure}[t!]
		\centering
		\includegraphics[width=\textwidth]{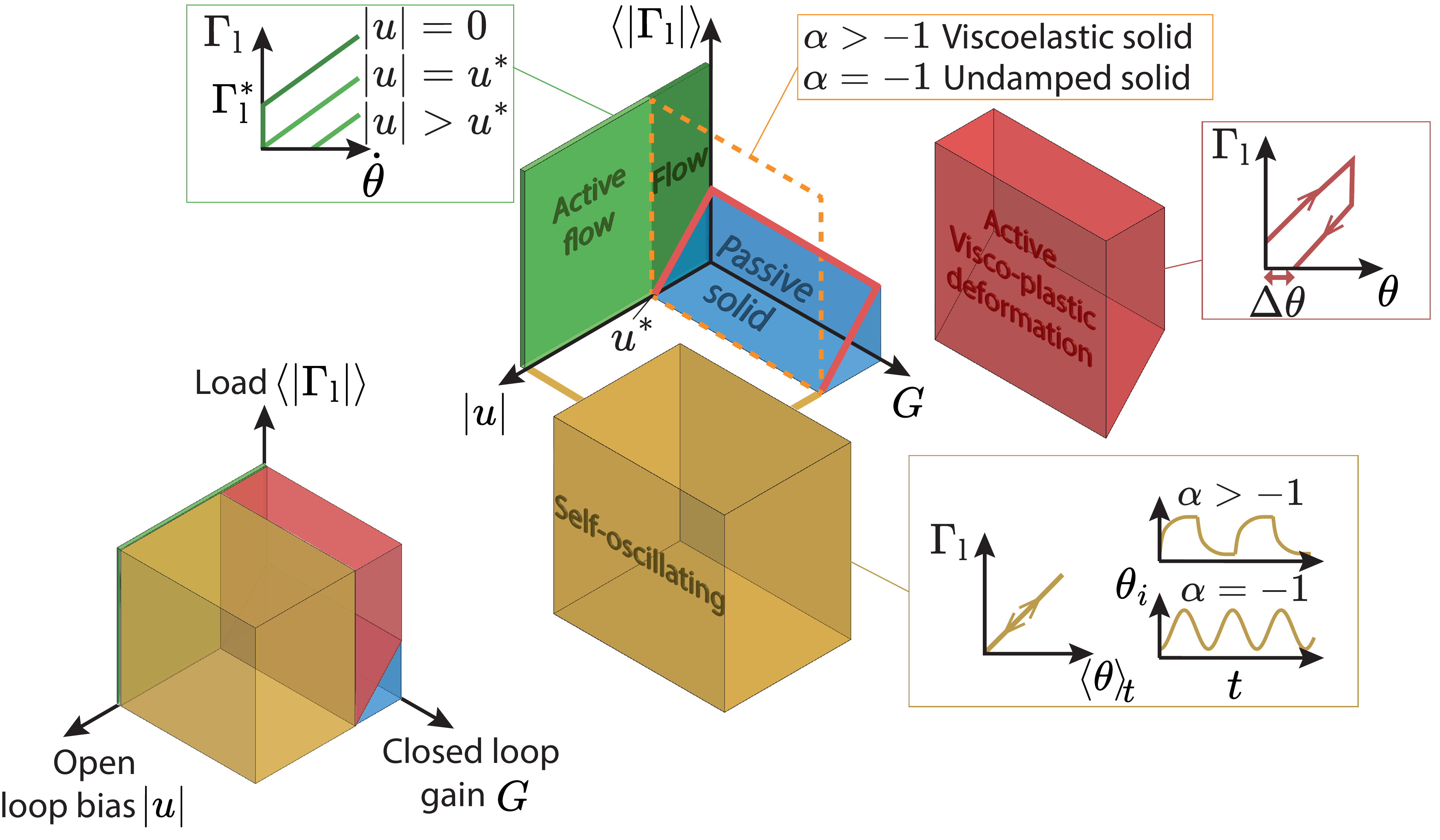}
		\caption{\textbf{Dynamical States of a Granulobot Aggregate.} Different states a Granulobot aggregate can exhibit as a function of open loop voltage bias $|u|$, closed loop proportional gain $G$, and average applied torque load $\langle|\Gamma_\mathrm{l}|\rangle$ (lower left). The exploded view (center) shows in more detail the parameter dependence of the different states. Schematic diagrams in the boxes indicate the associated load responses of individual units (for experimental data and modeling see Material and Methods). Measured mechanical behaviors of close-chain Granulobot aggregates are shown for  the active liquid-like state in Fig.~\ref{fig_4}C (relaxation time scale corresponding to an effective viscosity) and  Fig.~\ref{fig_5}A (strain-independent load response), for the visco-plastic solid state in Fig.~\ref{fig_5}A (hysteretic load response), and for the active, self-oscillating solid state in Fig.~\ref{fig_5}B (elastic load response).}
		\label{fig_3}
	\end{figure}
	
\begin{figure}
		\centering
		\includegraphics[width=\textwidth]{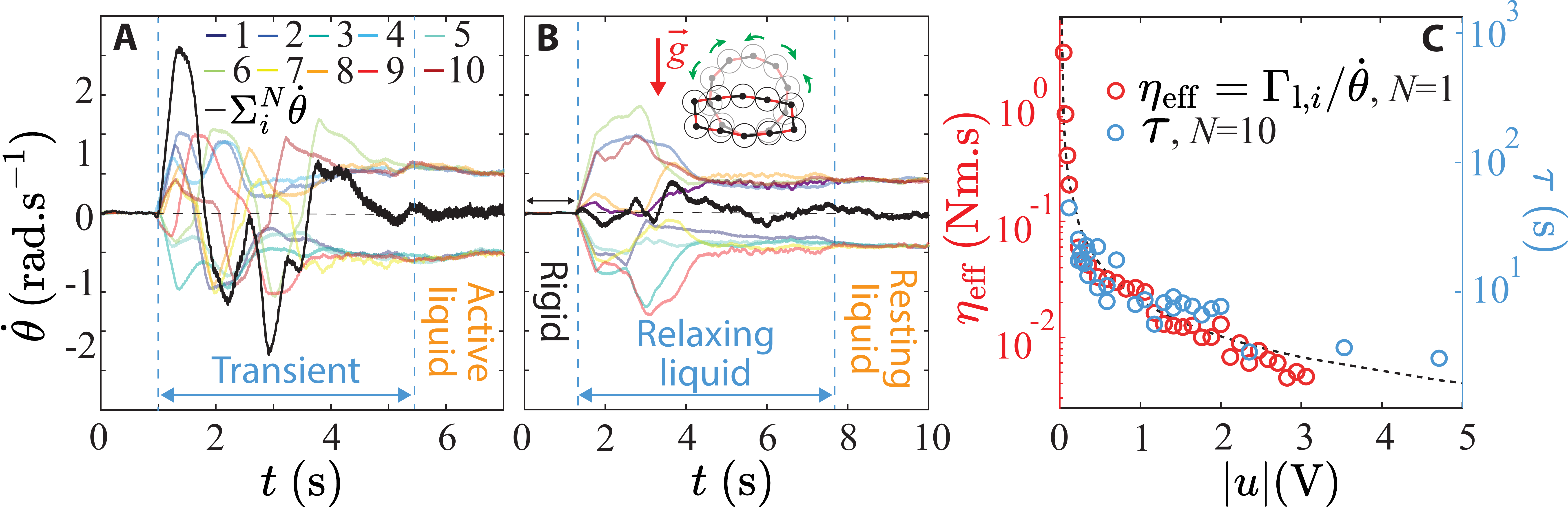}
		\caption{\textbf{Active Liquid-like state in a closed-chain aggregate.} $N=10$ units with $\alpha=0$ and $G=0$. (A) Rotation speeds of individual units as a function of time for an aggregate placed horizontally with $u_{i}=|u|\mathrm{sgn}(\dot{\theta})$. Here $|u|=1$~V. (B) Rotation speed of individual units as a function of time during a vertical collapse experiment with $|u|=1$~V. (C) Red circles: Evolution of a unit's damping parameter $\eta_{\mathrm{eff}}=\Gamma_{\mathrm{l},i}/\dot{\theta_i}=\eta_\mathrm{0}+k(u^*-|u|)\dot{\theta_i}^{-1}$, evaluated from measurements of $\dot{\theta_i}$ on a single unit, as a function of voltage bias $|u|$, for a fixed load $\Gamma_{\mathrm{l},i} =$\SI{0.020}{\newton\meter}, where $m$ is the unit's mass, $r$ its radius and $g$ the acceleration of gravity. Black dotted line: $\eta_{\mathrm{eff}}$ evaluated using the analytical form of $\dot{\theta_i}$ obtained from Eq. 3 (see Methods), and for a typical load $\Gamma_{\mathrm{l},i}=mgr$. Blue circles: Evolution of the relaxation time scale $\tau$, measured for the collapsing chain, as a function of voltage bias $|u|$.}
		\label{fig_4}
	\end{figure}

\begin{figure}
		\centering
		\includegraphics[width=\textwidth]{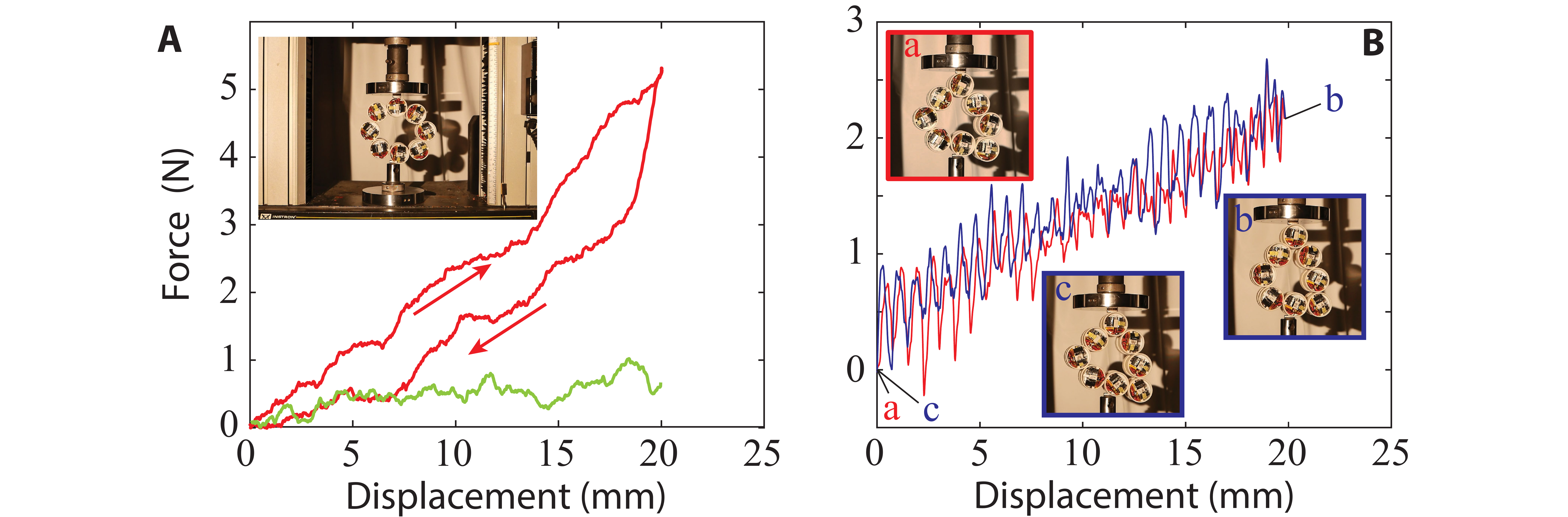}
		\caption{\textbf{Tensile load response of a closed-chain aggregate.} Force-displacement curves with $\alpha=-1$, $N=8$~units, measured at a displacement rate of \SI{1}{\milli\meter\per\second}. (A) Liquid-like response with $G=0$, $|u|=0.15$~V$\simeq u^*_0>u^*$ (green curve), and hysteretic response in the viscoplastic solid state with $G=$\SI{2.0}{\newton\meter\per\radian}, $|u|=0.1$~V$<u^*$. Inset: Still frame from a video of the Granulobot chain in the Instron materials tester used to generate the data in (A) and (B). In this image the back side of units is visible, showing the circuit board and the battery. (B) Reversible elastic response in the self-oscillating solid with $G=$\SI{2.0}{\newton\meter\per\radian} and $|u|=1$~V$>u^*$ (loading in red, unloading in blue). The periodic variations are due to shape oscillations of the aggregate. Insets: Enlarged images of the Granulobot chain in the Instron showing identical configurations at the beginning (a) and end (c) of the load cycle, and elongation when the maximum strain was applied (b).}
		\label{fig_5}
	\end{figure}

\begin{figure}
		\centering
		\includegraphics[width=\textwidth]{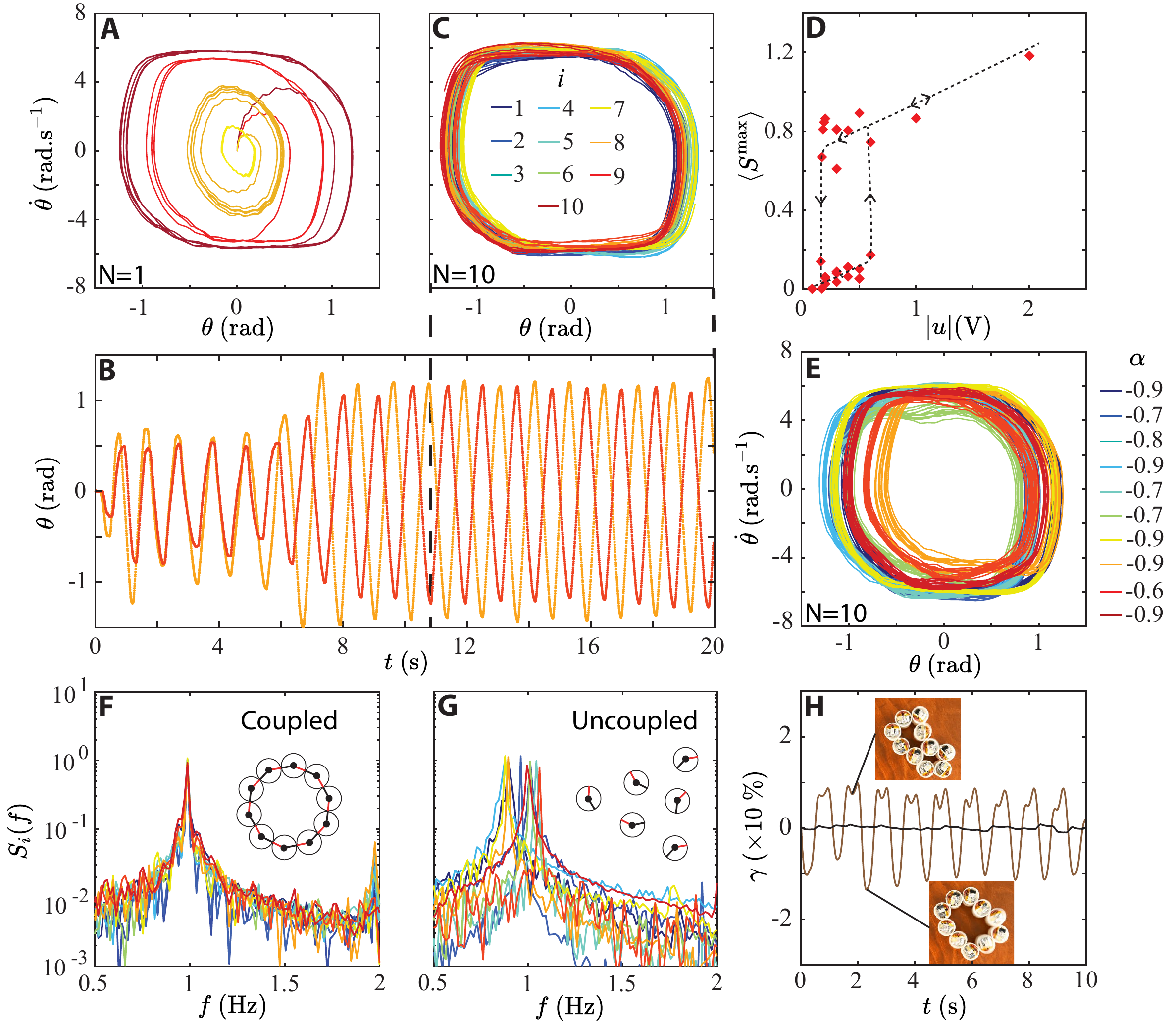}
		\caption{\textbf{Self-oscillating solid-like state in a closed-chain aggregate.} Granulobots are laid horizontally onto a slippery surface (Movie S3). $G=1.0$~Nm/rad and $u^*=0.11$~V. (A) Phase space $(\theta,\dot{\theta})$ of a single isolated unit for different $|u|$ and $\alpha$. $|u|=2$~V, $\alpha=-0.9$ (brown); $|u|=1$~V, $\alpha=-0.9$ (red); $|u|=1$~V, $\alpha=-0.5$ (orange); $|u|=1$~V, $\alpha=0$ (yellow). (B-H) $N=10$~units. (B) Angle as a function of time for two coupled neighboring units in the chain aggregate, after turning all units on at $t=0$. $|u|=1.0$~V. (C) Limit cycles followed by all coupled units in the steady regime of the experiments corresponding to (B). (D) Maximum of power spectrum averaged over $N=10$ units $\langle S^{\mathrm{max}} \rangle=\frac{1}{N}\sum{S_{i}^{\mathrm{max}}}$ as a function of $|u|$, where $S_{i}^{\mathrm{max}}$ is the maximum of the single-sided amplitude spectrum determined by applying FFT to $\theta(t)$, as in (F) and (G). Units are in the same aggregate configuration as in (B,C), with $\alpha=-0.9$. $|u|$ is gradually increased from $0$ to $2$~V, and decreased back to $0$~V. (E) Limit cycles observed in an aggregate with $\alpha \in [-0.9,-0.6]$, $|u|=1.0$~V. (F,G) Power spectrum observed for each of the units when coupled into a ring aggregate (F) and when isolated (G). (H) Shape deformation $\gamma=\frac{A-A(t=0)}{A(t=0)}$ over time, with $A$ the area covered by the assembly, for an aggregate with $\alpha_i=\alpha$ as in (B,C) (black), and with a range of $\alpha$ as in (E,F) (brown).}
		\label{fig_6}
	\end{figure}

	\paragraph*{\textbf{Granulobot Aggregate Behavior}\\}
	Figure~\ref{fig_3} shows the different states of a Granulobot aggregate as a function of average applied torque load $\langle | \Gamma_\textrm{l} | \rangle$ to the aggregate, and two of the control parameters, the voltage bias magnitude $|u|$ and the gain $G$ of the feedback for the angular rotor position. Each unit within the aggregate experiences a typical torque load of magnitude $\langle | \Gamma_\textrm{l}|\rangle$, while both $|u|$ and $G$ are set equal for all Granulobots (the sign of $\Gamma_\textrm{l,i}$ and $u_i$ can vary from unit to unit). The dissipation parameter $\alpha$ scales the effective viscosity and the relaxation time respectively in the liquid and the solid state of the aggregate, but for $\alpha \geq -1$ does not change the behaviors qualitatively. The load responses of individual units corresponding to the different aggregate states are sketched in the boxes (see Materials and Methods for experimental data and modeling). In the following parts we describe the different behaviors that corresponds to the states colored blue, green, red and yellow in Fig.~\ref{fig_3}, and provide experimental data for closed-chain aggregates of $N=8-10$ in Figs.~\ref{fig_4}, \ref{fig_5}, \ref{fig_6} (see Materials and Methods for discussion on variability among units and its effect on crossing between states).
	
	\emph{Passive State.} 
	
	If all biases $u_i = |u|$ are sufficiently small, the aggregate remains static until the frictional torque $\Gamma_\textrm{f}$ in a unit can be overcome. This is the passive state, shaded blue in Fig.~\ref{fig_3}. When contacts among the units in the aggregate involve only magnetic linkages, the passive aggregate has some degree of elasticity reflecting the stiffness of the magnetic coupling. When both magnetic and nonmagnetic contacts are involved, as in densely packed configurations where a given unit can also be in frictional contact by touching other units, this can create geometric constraints and lead to a granular jammed state with potentially very little elastic behavior (Fig \ref{fig_1}C). Such jammed state's response to load will depend strongly on the specific overall configuration of units within the aggregate.
	
	In the following, to simplify the presentation, we consider aggregates where all non-magnetic contacts slip easily and their contribution to friction can be neglected. In this case, the boundaries of the passive state in Fig.\ \ref{fig_3} are defined by the weakest coupling between units in the aggregate, which corresponds to Granulobots where the bias $u_i$ has the same sign as the load, thus effectively reducing the frictional resistance. This leads to a yield torque $\Gamma^*_\textrm{l} = \Gamma_\textrm{f}-k|u|$, which vanishes once $|u|$ exceeds $u^* = \Gamma_\textrm{f}/k$, and gives rise to the wedge shape of the passive state in Fig. 3.
	
	\emph{Liquid-Like States.} 
	
	These states occupy the green $(|u|,\Gamma_\textrm{l})$ plane in Fig.~\ref{fig_3}, where $G = 0$.
	The bias voltage $u^*$ delineates behavior akin to a yield stress fluid for $u < u^*$ and corresponding to an active fluid for $u > u^*$. For a highly deformable, steady liquid-like load response to emerge, neighboring magnetically coupled units ideally all need to rotate in opposite directions. In an aggregate comprising a closed chain of an even number of units $N$, this translates into $\sum_i^N\dot{\theta_i} \approx 0$. Such regime is reached autonomously by setting $u_{i}= |u|\,\mathrm{sgn}(\dot{\theta_i})$ in each of the individual units (in this case onboard sensing of the rotor's direction of rotation is required). The mechanical coupling then drives a self-organization process that, after a brief transient regime, converges to a state with bias voltages of opposite sign on neighboring units, $u_{i}=-u_{i-1}=(-1)^{i}|u|$ (Fig.\ \ref{fig_4}A, Movie S2 part 1). Alternatively, the liquid state can be achieved in fully open loop, with no sensing of $\mathrm{sgn}(\dot{\theta_i})$, by initializing neighboring units via a global command to use $|u|$ of opposite signs (See Methods), and with $\alpha=0$. From Eq. 3, for $G = 0$ the load response of all individual units is described by an equation of the form
	\begin{equation}
		\Gamma_{\textrm{l},i} = (1 + \alpha)\eta_{\mathrm{0}} \dot{\theta}_i + k(u^* - |u|),
	\end{equation}
	written here for positive $\Gamma_{\textrm{l},i}$ and therefore $\dot{\theta}_i > 0$ and sgn($\dot{\theta}_i)$ = 1.
	
	If $|u| < u^*$, this relationship between applied load and resulting rotation speed has the same functional form as the relationship between applied stress and shear rate for a yield stress fluid (see sketch of $\Gamma_\textrm{l}(\dot{\theta})$ in Fig. 3). For such fluid an effective, rate-dependent viscosity $\eta_{\mathrm{eff}}$ is defined by the ratio of stress to shear rate. By analogy we here define a unit's viscous damping as $\eta_{\mathrm{eff}} = \Gamma_{\textrm{l},i} / \dot{\theta_i}$.
	
	To compare this with the viscous behavior of a multi-unit Granulobot aggregate in its liquid state, we characterize the collective viscous flow by performing experiments where we time the collapse of a  circular chain units (Fig.\ \ref{fig_4}B, Movie S2 part 2). We choose an initial configuration such that before the collapse, for $u=0$, the  load for all units in the aggregate is slightly below their frictional yield  torque and therefore the chain's circular shape is maintained. We then switch on the bias $|u|$ to drive the system into the liquid state and measure the relaxation time $\tau$ needed for the aggregate to reach its minimum potential energy, i.e., its equilibrium state. To bypass the transient phase of the self-organizing process and ensure the system behaves as a liquid as soon as the voltage bias is applied, we enforce $u_{i}=(-1)^{i}|u|$. 
	Note how adjacent units rotate in opposite directions and how this complex sequence of interdependent rotations emerges simply in response to external forcing. As such, it can accommodate arbitrary starting configurations and will always proceed in a liquid-like manner toward the equilibrium configuration that minimizes gravitational potential energy. 
	
	Figure~\ref{fig_4}C shows measurements of $\eta_{\mathrm{eff}}$ (red circles) and $\tau$ (blue circles) as a function of the voltage bias $|u|$. The plotted $\eta_{\mathrm{eff}}$ is for an individual unit, obtained by applying a fixed torque $\Gamma_{\textrm{l},i}$ and measuring the resulting angular speed $\dot{\theta_i}$ as a function of $|u|$. For comparison with the collapse experiments, the torque load is chosen slightly lower than the friction torque, such that the rotor is also static at $u=0$. The experimental data are well described by an analytical form of $\eta_{\mathrm{eff}}= \Gamma_{\textrm{l},i} / \dot{\theta_i}$ using Eq.\ 4 to obtain $\dot{\theta_i}$ (black dotted line). We observe that the aggregate collapse time $\tau$ reproduces this behavior, consistent with a collective viscosity that scales with bias $|u|$ in the same way as $\eta_{\mathrm{eff}}$. This demonstrates how the collective viscosity can be tuned by setting the viscous load response of the individual units via $|u|$. Changing the damping via $\alpha$, which is set to zero in all data shown in this section, only scales the behavior with a prefactor~\cite{footnote_1}. 
	
	For vanishing voltage bias $|u|$, the passive state is approached and $\tau$ and $\eta_{\mathrm{eff}}$ diverge. Conversely, as $|u|$ is increased to the point that it reaches $u^*$, the yield stress vanishes and the aggregate viscosity attains a load-independent value that scales as $\eta_{\mathrm{eff}}\sim(1 + \alpha)\eta_0$. Effectively, the system now behaves like a Newtonian liquid. If $|u|$ is increased above $u^*$, individual units rotate with finite speed $\dot{\theta_i} = \frac{k(|u| - u^*)}{(1 + \alpha)\eta_0}$ already in the absence of any significant load (see e.g. the final equilibrium state in Fig. \ref{fig_4}B and Movie S2). In this state the aggregate behaves in a manner that can be characterized as an active liquid with low effective viscosity. The green curve in Fig. \ref{fig_5}A presents data from a constant speed tensile test when $|u| = u^*$, i.e. when $\Gamma_f$ is actively compensated for by the voltage bias, so the rotor is still static when not loaded. We observe a quasi-flat, i.e. strain independent behavior, consistent with what we expect for a Newtonian liquid-like response. In practice, a unit starts to rotate at a slightly larger bias than the estimate from calibration torque-displacement curves $u^*_0>u^*=\Gamma_f/k$. This can be explained by a static friction at zero angular speed that is larger than the kinetic friction producing $\Gamma_f$ (see Fig.~\ref{fig_8}D in Materials and Methods).
	
	\emph{Solid-Like States}. 
	
	These states use a feedback to sense the rotor movement and drive it toward its equilibrium angular position during a load perturbation. They are colored red and yellow in Fig.\ \ref{fig_3}. With increasing $G$, the response of individual units, and thus also of the aggregate, becomes more elastic and rigid. 
	
	Considering a bias that always acts against the frictional torques $u_i=|u|~\mathrm{sgn}(\dot{\theta_i})$, Equation \ref{SteadyState} can be written as
	\begin{equation}
		\Gamma_{\textrm{l},i} =G\theta + (1 + \alpha)\eta_{\mathrm{0}} \dot{\theta}_i + k(u^* - |u|)\mathrm{sgn}(\dot{\theta_i}),
	\end{equation}
	
	When $|u| = u^*$ (vertical dotted plane in Fig.\ \ref{fig_3}), this behavior corresponds to a viscoelastic solid, the response of a damper and a spring in parallel \cite{Oswald}. In this specific state, vanishing damping, i.e. $\alpha=-1$, corresponds to an undamped mass-spring oscillator.  
	
	When $|u| < u^*$, the unit's response is analogous to a viscoplastic solid (colored red in Fig.~\ref{fig_3}), which corresponds to a spring, a damper and a frictional element in parallel \cite{Oswald}. In this regime, where the characteristic response time of a rotor is $\tau = (1+\alpha)\eta_0/G$, effective position control with a fast response is achieved by choosing a positive $\alpha$ and a large $G$. Otherwise, if $\alpha = -1$, the damping $\eta_{\mathrm{0}} \dot{\theta}_i$ is actively compensated for and a perturbation of the rotor that overcomes $\Gamma_\textrm{f}$ leads to steady undamped oscillations. As the biases will not completely cancel the frictional torques in individual units, if $G$ is not large enough, deformation will exhibit significant hysteresis $\Delta\theta=k(u^* - |u|)/G$. As shown in the tensile test data presented in Fig. \ref{fig_5}A, the overall aggregate will then also respond to load with some hysteresis, i.e. plasticity.
	
	If we choose biases $|u_i| >u^*$, the response of the aggregate to a quasi-static tensile test no longer is hysteretic but becomes reversible, as shown in Fig. \ref{fig_5}B. This is the signature of an elastic solid. However, unlike an ordinary elastic solid, in this state (colored yellow in Fig.~\ref{fig_3}), the internal dynamics of the aggregate is more complex: each rotor's unit is driven away from its equilibrium angle and oscillations arise spontaneously, even without applied load. The units are driven by the bias to "self-oscillate" such that, when isolated, they exhibit well defined limit cycles with amplitude and frequency fixed by $|u|$, $G$ and $\alpha$ (Fig.\ \ref{fig_6}A). As individual units' rotors are now subject to inertia (see Material and Methods for modeling), which manifests itself through spring-like oscillation, they can transfer momentum to each other when magnetically coupled in an aggregate, and thus perturb their neighbor's dynamical state. In a chain of units, after an initial transient, such purely mechanical coupling can lead to the self-organization of collective oscillatory states with well defined global limit cycles that remain unchanged over time. Specifically, in a closed loop of an even number of units that all have the same parameters, and that remain unperturbed, we observe a synchronization process (see Movie S3, part 1). Figure \ref{fig_6}B displays such situation in an aggregate of $N=10$ units, by showing the angles of two consecutive units as a function of time, right after turning them on. After a transient, units lock in their phase and amplitude. The frequency and amplitude observed in the aggregate are the same as for an isolated unit (Fig.\ \ref{fig_6}C). 
	
	Moreover, the transition between a disordered dynamics to a synchronized behavior is strongly hysteretic  (Fig.~\ref{fig_6}D), with long transient regimes at biases close to $u^*$, and almost instantaneous transitions at larger biases. This hysteresis implies that, for a certain range of bias values $|u|$, the Granulobot aggregate can be in two states, with and without limit cycle. With a sensitivity that can be tuned with the voltage bias within this range, this means that switching between these behaviors can be triggered by mechanical perturbations from the environment the robot interacts with, without electronic feedback.
	
	Remarkably, when the system synchronizes, consecutive units' phases alternate between $0$ and $\pi$, which corresponds to $\dot{\theta}_i=-\dot{\theta}_{i+1}$, and $\sum^N_i\dot{\theta_i}\sim 0$ (Fig.~S2B). In this situation, the contour of the chain remains static in time with no deformation. We label this collective steady state a self-oscillating solid (a similar phenomenology is observed when choosing $u_i$ across all units with the same sign, regardless of $\Gamma_\textrm{f}$~\cite{footnote_2}). In this state, perturbations that are too small to switch the aggregate to disordered dynamics can still affect its behavior. For example, placing a previously synchronized closed-chain loop upright in a gravity field when performing a tensile test leads to small global periodic deformations of the structure that can be seen by the oscillations in the force signal in Fig.~\ref{fig_5}B. But remarkably, on average the response to loading remains that of a purely elastic solid, with collective self-organized properties that ensure continuous deformation across all of the aggregate (see Movie S3, part 2).    
	
	Furthermore, such global shape oscillation can also be triggered internally when units are assigned different parameters and thus exhibit different self-oscillation properties. In such setting, we still observe collective limit cycles characterized by self-organization with frequency locking. Figure~\ref{fig_6}E shows such collective limit cycles for an assembly with a distribution of $\alpha$ in the range $[-0.9, -0.6]$. Figures \ref{fig_6}(F,G) compare the frequency spectra of each unit coupled in an aggregate with those for isolated units. We see that the mechanical coupling selects a single collective frequency despite the spread in frequencies of isolated units, with phases that are locked in time. Meanwhile, there remains a spread in oscillation amplitudes. To accommodate this, the phases between consecutive units in a chain are shifted (See Fig. S2) and the aggregate's contour deforms globally and periodically (see Movie S3 part 3). Figure~\ref{fig_6}H shows the evolution of such deformation as a function of time for an assembly with the same range of $\alpha$ as in Figure~\ref{fig_6}(E,F) and compares it with the observation for an assembly with identical parameters among all units, as in Fig.~\ref{fig_6}(B,C). With variable damping, the units no longer synchronize and a wave propagates instead.
	
	\paragraph*{\textbf{From collective states to decentralized locomotion tasks}\\}
	
	\begin{figure}[h!]
		\centering
		\includegraphics[width=\textwidth]{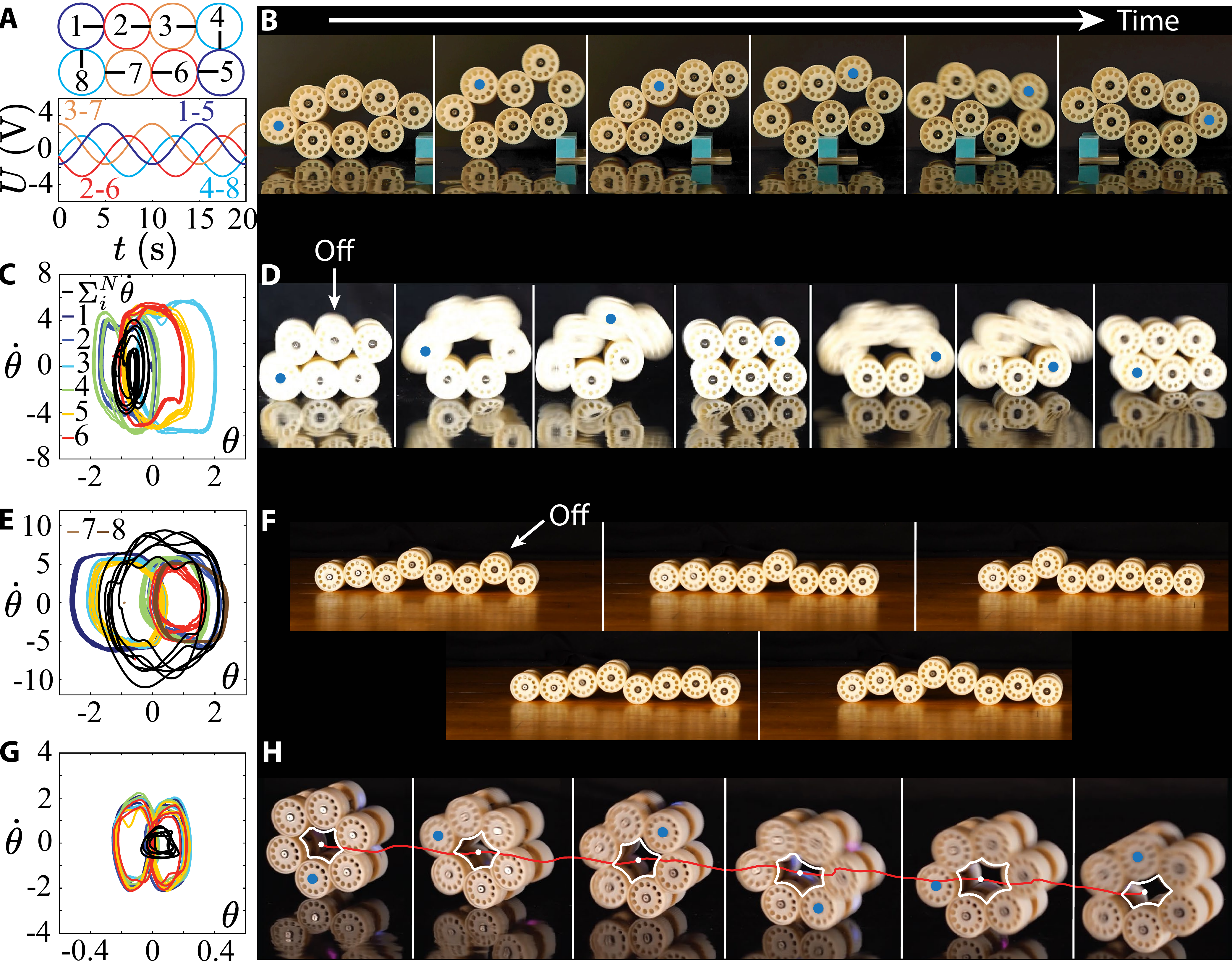}
		\caption{\textbf{Exemplary Granulobot locomotion strategies.} (A) Liquid-like gait using a set of voltage biases with shifted phases in an aggregate of $N=8$ units (see Materials and Methods, $u_\mathrm{0}=0.7$~V and $A=2.4$~V). (B) Snapshots of a Granulobot aggregate using the gait (A) to move over an obstacle. (C) Limit cycle reached by a closed chain aggregate in the self-oscillating solid state with $N=6$ units, after turning  unit $i=1$ off ($|u|=2$~V, $\alpha=-0.9$ and $G=1.0$~Nm/rad). (D) Snapshots of a steadily moving self-organized aggregate corresponding to the gait in (C). (E) Limit cycle of an open chain aggregate in the self-oscillating solid state with $N=8$ units, after turning unit $i=7$ off ($|u|=1$~V, $\alpha=-0.9$ and $G=0.5$~Nm/rad). (F) Snapshots of a steadily moving self-organized aggregate corresponding to the gait in (E). (G) Limit cycle from a leader-follower scheme to implement a propagating perturbation (inertial shape-shifter). Choosing $|u|$=0, $G=10.0$~Nm/rad and $\alpha=1.0$, units can be controlled in  displacement (large G with $|u|<u^*$), enabling fast collective motion with small deformation by turning into a ``pumping" wheel. (H) Snapshots corresponding to gait (G). The blue dot corresponds to unit $i=1$. In (C), (E) and (G) the units of $\theta$ and $\dot{\theta}$ are rad and rad/s, respectively.} 
		\label{fig_7}
	\end{figure}
	
	The behaviors summed up in Fig.~\ref{fig_3} show a wide range of responses to external loads and constraints. Such mechanical feedback from the environment can be seen as an emergent and decentralized collective mechanical sensing ability that can be harnessed for control. These collective states can be driven out of their effective equilibrium to produce dynamical shape shifting solely by preventing individual units behavioral state, enabling decentralized deformation based tasks. Furthermore, the use of voltage control via a single functional form (Eq. 2) allows to continuously transition between  states of different response type and switch autonomously between strategies depending on environmental or computation constraints. We demonstrate this ability by implementing three different locomotion gaits. Using the liquid-like, the self-oscillating and the viscoplastic mechanical responses, a robot can move over an obstacle without sensors, self-organize into a steadily translating aggregate, either on frictional or slippery surfaces, solely via mechanical coupling, and can move quickly and efficiently using its own inertia when it maintains overall rigidity.  
	
	\emph{Morphing Liquid for Sensorless Obstacle Management.} 
	We leverage the aggregate's ability to change its rigidity and behave as a liquid to locomote over an obstacle,  using only mechanical feedback from the environment. For an aggregate to deform liquid-like while translating its center of mass  without sensor readout, we must find a form of $u_i$ that implements a gait while  satisfying $\sum^N_i\dot{\theta}\sim0$. Here we show that this can be achieved with $\alpha=G=0$ by using the functional form for the individual $u_i$ displayed in Fig.\ \ref{fig_7}A, which we have inferred from a reverse kinetic approach (see Materials and Methods). This gait relies on synchronous periodic deformation that we can implement with both closed or open chain configurations (see Materials and Methods), by allowing neighboring units to wirelessly synch their clocks and correct phase drift over time. Figure~\ref{fig_7}B shows a snapshot of an aggregate "flowing" over an obstacle to pass it (see Movie S1). 
	
	\emph{Undulator: Emergent aggregate gait with coupled oscillators.}
	Figure \ref{fig_6}H shows that a spread in the units' oscillation properties in an aggregate leads to limit cycles that generate periodic collective  shape deformations. Even when there is no spread, similar collective shape deformations can be triggered by mechanical interaction with the environment. Specifically, when an aggregate of synchronized units, all with equal control parameters, is placed vertically onto a non-slip surface, limit cycles remain, but small regular left-right periodic deformations emerge due to frictional constraints: units that are in contact with the non-slip surface cannot rotate as much as the others. In such setting, perturbing this state further by turning off one unit modifies the limit cycle (Fig.\ \ref{fig_7}C), breaks the left-right symmetry of the collective periodic deformation, and amplifies it. This can lead to a sustained locomotion of the aggregate (Fig.\ \ref{fig_7}D) without the use of wireless communication, solely relying on unit-unit mechanical interactions, and hence minimal computation (See Movie S3 part 4). To locomote with a closed chain, the left-right symmetry of the periodic deformation can also be broken by using an odd number of units with identical oscillation parameters (see Movie S1). In an open chain configuration, periodic deformation can be used to locomote on a slippery surface. In that case, perturbing the properties of an oscillator in one portion of the chain can trigger an undulating deformation that alternates contact with the ground periodically between different sections of the chain (Fig.\ \ref{fig_7}(E,F), Movie S1). 
	
	\emph{Inertial Shape-Shifter.} 
	Here we demonstrate the implementation of a local displacement control approach to generate fast and efficient rolling gaits in rigid solid aggregates. This is achieved via fast and overdamped local feedback loops in the viscoplastic region of Fig.~\ref{fig_3}, with large $G$, $|u|=0$ and an optimal $\alpha$. Small, appropriately timed shape perturbations are propagated through a ring-like chain of coupled units to generate coordinated locomotion that leverages the assembly's overall inertia. The shape perturbations are enforced by controlling the rotors' absolute angles as a function of time (see Methods for details). The aggregate's locomotion is triggered in a decentralized way by sending the same message once to all units. Neighbor-neighbor communication and a leader-follower algorithm then implement a predefined angle perturbation that propagates through the chain. Depending on the algorithm parameters, the aggregate can either crawl or roll, the latter demonstrating the faster and most energy efficient locomotion strategy as only small rotor displacements are required (Fig.~\ref{fig_7}(G,H) and Movie S1).
	
	\section*{Conclusions}
	
	The Granulobot represents a novel robotic system inspired by granular material where motorized units can flexibly and reversibly couple with each other and self-assemble into larger aggregates. A minimal design with a single activated degree of freedom per unit enables aggregates to reconfigure dynamically and to transition between solid- and liquid-like responses under gravity. In contrast with modular robotic systems that require successive steps of detachment and reattachment of individual units to change their shape, a Granulobot aggregate can deform by continuous local displacement, similar to a deforming soft material. Since such deformation is a function of the Granulobots' dynamical states, this results in the wide range of collective mechanical behaviors shown in Fig.\ \ref{fig_3}. Furthermore, as these different response properties are all encoded in the functional form of the control voltage $U_i$, this allows to transition continuously between liquid-like behavior associated with open loop control, and solid-like behavior where an onboard feedback loop monitors the local rotor displacement sensor. 
	
	The properties associated with several of the Granulobot aggregate states make it possible to use mechanical interaction with the environment to implement decentralized locomotion strategies without the use of electronic environmental sensing and at minimal computation cost. In the liquid-like state, this allows an aggregate to pass over an obstacle by "flowing" over it. In the self-oscillating solid state, Granulobots self-organize strictly via mechanical coupling into well-defined limit cycles. In this setting, changes in the control parameters, as well as in the mechanical environment where robots operates in, can trigger periodic deformations. Using this feature, aggregates can self-coordinate periodic shape changes to slide on a slippery surface and roll on a no-slip surface, thus leveraging physical interactions with their environment to generate  locomotion \cite{Ijspeert2021}. Furthermore, the bistability we observe in Granulobot aggregates for a range of bias voltages enables a minimal, purely mechanical `decision-making': sufficiently large external perturbations can switch the behavior of the system between disordered and self-organized states without requiring electronic sensing. Reminiscent of mechanisms found in living systems \cite{Sridhar2021}, this opens up a promising perspective in distributed robotics to develop collective behavioral capabilities.
	
	The viscoplastic state makes it possible to maintain an aggregate shape and exhibits less sensitivity to perturbation compared to the self-oscillating solid. This state is more amenable to the implementation of traditional displacement control, where parameters can be tuned for a rigid response that is insensitive to external environmental loading, similar to controlling a robotic arm. In our experiments with closed-chain aggregates, this is advantageous when implementing fast locomotion using inertia. It also demonstrates the ability of the design to combine in a decentralized way our physical approach with more widely used kinematic planned strategies. This is also the regime in which controlled neighbor reconfiguration can be implemented in a usage more similar to existing modular robotic system. But such reconfiguration can happen spontaneously in any of the accessible states through collective mechanisms.
	
	We view the different locomotion strategies explored here as first examples that demonstrate the versatility of the Granulobot platform. As such, we anticipate that many additional locomotion strategies could be found as well as possible new tasks. A promising way to explore them could be to implement machine learning approaches such as in \cite{Bongard_2006,Lipson2014}. A more ambitious but crucial step would be to implement learning in a fully decentralized way, which could allow aggregates to select autonomously strategies that are relevant to the environment and constraints they encounter. In this context, conventional machine learning approaches are conceptually challenging to apply since error functions are ultimately estimated centrally. We believe that the Granulobot platform offers a promising perspective for tackling this problem by combining the collective behavior emerging from mechanical interactions among the units with local error optimization executed in the units' microcontrollers.
	
	Furthermore, more complex packing and coupling structures could be created. This could be achieved by increasing each unit's connectivity with additional freely rotating magnets and, in dense packing configurations, exploiting the metastability associated with granular jamming \cite{Jaeger2015}. For instance, it should be possible to maintain structural rigidity at no energy cost and switch configurations by actuating only few units where resistance to reconfiguration is weaker. In such setting, larger connectivities would increase the effect of couplings and could allow for more collective mechanical adaptability \cite{Peleg2018}. Additionally, while the Granulobot aggregates discussed in this work are effectively two-dimensional, arranging different aggregates next to each other with elements that connect them suggests a straightforward extension to a 3D version. Finally, the simple contact and actuation principle that we propose could be integrated in larger scale programmable matter \cite{Piranda2018}.
	
	\section*{Materials and Methods}
	
\begin{figure}[t!]
		\centering
		\includegraphics[width=\textwidth]{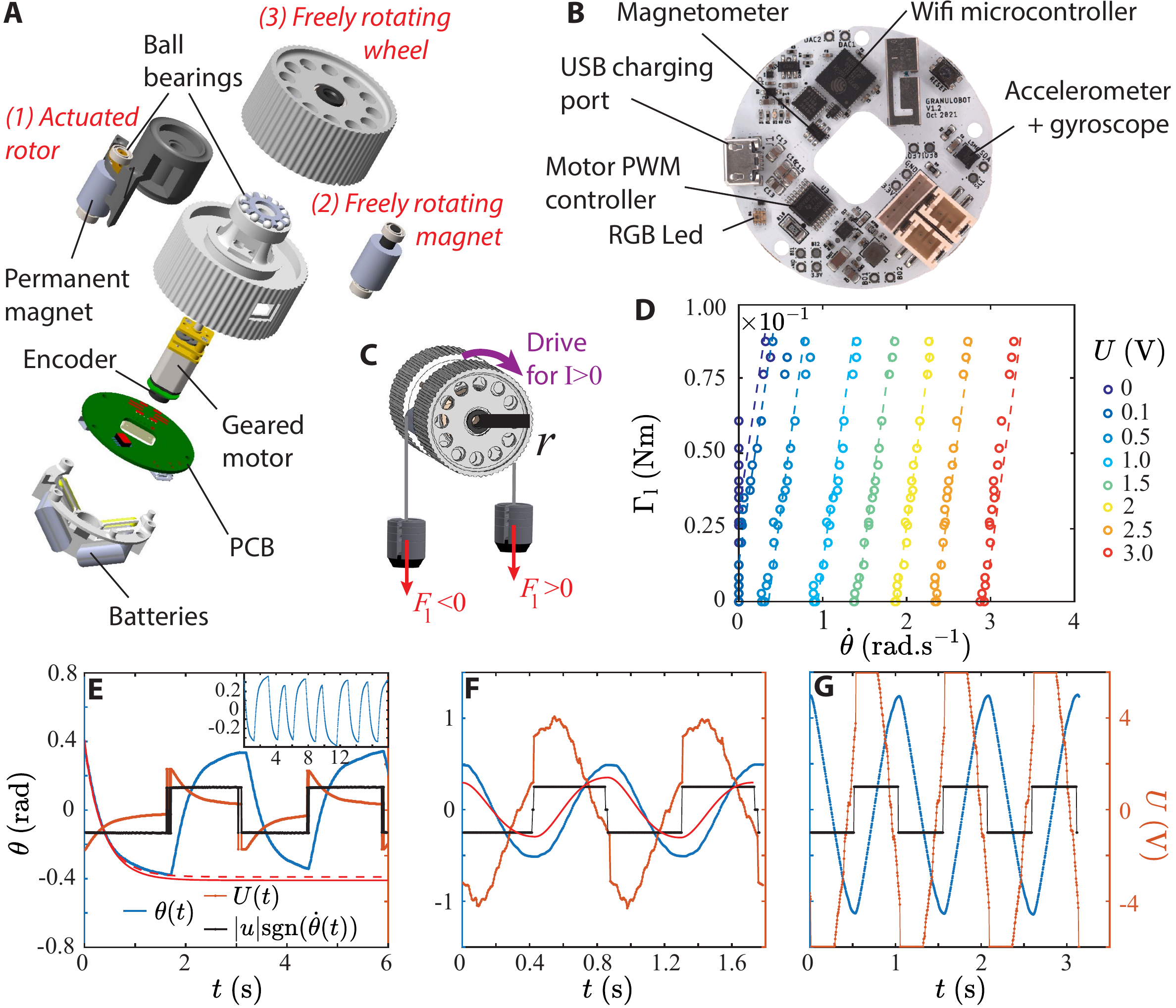}
		\caption{\textbf{Granulobot design, rotor response to load, and actuation.} (A) Exploded view of  Granulobot 3D model. (B) Assembled electronic circuitry, custom-designed for driving and real-time measurements. (C) Sketch to indicate sign convention. (D) Torque load as a function of rotation speed for different voltages. Dotted lines are fits to Eq. \ref{SteadyMotion} with fixed parameters $R=9.6~\Omega$, $K_\textrm{e}=2.65\times 10^{-3}$~V.s.rad$^{-1}$. $K_\textrm{t}=5.8\times 10^{-3}$~Nm$A^{-1}$ and $\Gamma_f=0.024$~Nm are determined by averaging fit values obtained for each voltage tested. $\beta=379$. (E-G) $\theta$ (blue), voltage bias $u$ (black) and total voltage $U$ produced by the circuitry (orange) as a function of time in a self-oscillating Granulobot with $\Gamma_\textrm{l}=0$. Inertia of rotor with magnet $I_{\theta} =0.036~\textrm{kg} \times (0.020~\textrm{m})^2$, $k=0.229$~Nm.V$^{-1}$, and $\eta_0=0.231$~Nm.s.rad$^{-1}$ (see Suppl. Materials). (E) Overdamped self-oscillation ($\zeta>1$). $\alpha=0$, $|u|=1$~V, $G=0.5$~Nm/rad, corresponding to $\zeta=1.37$ and  $\omega_0=5.88$~rad.s$^{-1}$. Red curves are Eq. 11 for $\Gamma_f=0.024$~Nm (solid) and $\Gamma_f=0.034$~Nm (dashed). (F) Self-oscillation in the crossover regime ($0<\zeta<1$). $\alpha=-0.5$, $|u|=1$~V, $G=1$~Nm/rad, corresponding to $\zeta=0.48$ and $\omega_0=8.31$~rad.s$^{-1}$. Red curve is the prediction from Eq.~\ref{sol_2}, leading to  $\omega=7.29$~rad.s$^{-1}$ and amplitude $\theta_n=0.30$~rad. (G) Underdamped self-oscillations ($\zeta=0$). $\alpha=-1$, $|u|=1$~V, $G=1$~Nm/rad, corresponding to $\omega_0=8.31$~rad.s$^{-1}$.}
		\label{fig_8}
	\end{figure}

	\paragraph*{\textbf{Robotic prototype}\\}
	The Granulobot prototype discussed in this work consists of 3D printed parts assembled with ball bearings, permanent magnets, a custom battery-powered electronic circuit and a geared DC motor (Fig.\ \ref{fig_8}A). Each individual Granulobot unit is comprised of three co-axial components that can rotate independently from each other: (1) a first wheel, which contains the DC motor (N20 from Pololu) that drives a rotor with a magnet at its end (referred as actuated magnet in the main text); (2) a second rotor, also with a magnet at its end, that can move freely around the central motor axis (referred as free magnet in the main text); and (3) a freely rotating second wheel (shown at the top of the exploded view in Fig. 6A) that helps align  the Granulobot units when making contact and stabilizes an assembled aggregate against tipping over. Both rotors hold hollow cylindrical neodymium magnets (N54 from Supermagnetman, size 3x12x15.5~mm$^3$) via a shaft with ball bearings on each end, which let the magnets rotate freely. The active rotor attaches to the motor shaft with two screws to adjust parallelism. The motor's gear train can be changed to different ratios $\beta$ depending on the application, with values ranging commercially from 5:1 to 1000:1. Additional freely rotating magnets can be added to increase the connectivity of a single unit and build more complex aggregate structures if desired (see Fig. S1).
	
	Individual Granulobot units have an overall width $l=62$~mm and a wheel diameter $2r=48$~mm (OD). 3D printed parts are made with a UV-curing polyjet printer (Stratasys J850) using an ABS-like resin (RGD511 and 535) and with a Phrozen Mini 4k printer using Loctite Onyx 410 engineering resin. 
	The total weight of a single Granulobot unit with geared motor, printed circuit board (PCB), ball bearings, magnets, and two $220$~mAh batteries is $98$~g. 
	
	The electronic circuitry consists of a Wi-Fi enabled microcontroller (Espressif ESP-32), an H-bridge to control the motor, a power management circuit with 3.3 V and 6 V regulation from two single-cell lithium-ion polymer (Li-po) batteries, a 6-axis linear and gyroscopic accelerometer, a magnetic sensor with sensitivity in the range of the field produced by the rotor magnets, and a continuous magnetic encoder to measure the motor rotation. All electronic components are assembled on a printed circuit board designed to fit within the circular Granulobot body (see Fig.~\ref{fig_8}B). The motor voltage $U$ is produced by two microcontroller digital outputs using pulse width modulation (PWM), averaged out by the low frequency response of the motor system (See Supplementary Materials for PWM calibration). The entire circuit is controlled and monitored by a firmware coded in C++ with Arduino and Esp32 libraries. Data can be sent and received by Wi-Fi between individual  units as well as with an external computer. For the latter, we developed Python software that gathers data for post-analysis and visualisation, at a rate of approximately $200~$Hz, and that can execute scripts with a sequence of instructions. This interface also allows to reprogram remotely all robot units simultaneously for firmware changes.

	\paragraph*{\textbf{Steady-State load response of individual Granulobot units}\\}
	
	In a DC motor the current $I$ through the winding generates a torque $\Gamma=K_\textrm{t}I$, where the constant $K_\textrm{t}$ depends on the mechanical and electromagnetic properties of the motor \cite{Hughes1990}.
	Our sign convention is that positive current $I$ produces positive torque drive and positive gearbox output rotor speed $\dot{\theta}$ in the absence of any load. In Fig.\ \ref{fig_8}C positive torques correspond to clockwise rotation. The torque balance in the steady state of a loaded motor can then be expressed as
	\begin{equation}
		\beta K_\textrm{t}I+\Gamma_l-\Gamma_f \mathrm{sgn}(\dot{\theta})=0,
		\label{Tbalance}
	\end{equation}
	where $\beta$ is the gear box ratio,  $\Gamma_l$ is the external load  (which can be positive or negative) on the motor's gearbox output, and $\Gamma_f >0$ is the magnitude of the friction torque that arises from the mechanical parts contacting each other inside the gearbox and the rotor system. 
	The sign function sgn($\dot{\theta}$) = $\dot{\theta}/|\dot{\theta}|$ appears because the friction torque always opposes any rotation. Using Kirchhoff's law, the motor current $I$ during steady rotation obeys
	\begin{equation}
		RI = U + E,
		\label{backEMF}
	\end{equation}
	where $R$ is the motor resistance, $U$  the  voltage applied to the motor, and $E$  the back electromotive force (e.m.f.). The latter opposes the motion of the inner rotor, which spins at speed $\omega=\beta \dot{\theta}$, and can be expressed via Faraday's law as $E= - K_\textrm{e}\omega$, with $K_\textrm{e}$ being the motor's EMF constant. Equation \ref{backEMF} combined with Eq.\ \ref{Tbalance} can be rewritten as:
	\begin{equation}
		\beta\frac{ K_\textrm{t}}{R}U +\Gamma_\textrm{l} - \Gamma_\textrm{f}\mathrm{sgn}(\dot{\theta})=\frac{ K_\textrm{t} K_e}{R}\beta^2\dot{\theta}. 
		\label{SteadyMotion}
	\end{equation}
	
	Theoretically $K_\textrm{t}=K_\textrm{e}$, but in practice these two value can be slightly different, and we  therefore  distinguish them in what follows. 
	Furthermore, $K_\textrm{t}$ tends to decrease with an increasing gear ratio; thus, it is important to estimate it whenever $\beta$ is changed (see Supplementary Materials for estimates of $K_\textrm{e}$ and $K_\textrm{t}$).
	Defining $\eta_\textrm{0} = \beta^2 K_\textrm{t} K_\textrm{e}/R$ and $k = \beta K_\textrm{t}/R$ we arrive at Eq. 1 in the main text for the case of steady-state rotation, i.e. when the frictional torque has been exceeded $|\Gamma_\textrm{l} + kU| > \Gamma_\textrm{f}$.
	
	From Eq.\ \ref{SteadyMotion} or, equivalently, Eq.\ 1 we expect the speed-torque curves to be linear with slope $\eta_\textrm{0}$ once the load $\Gamma_\textrm{l}$ exceeds the yield threshold torque. 
	When $kU$ becomes larger than $\Gamma_\textrm{f}\mathrm{sgn}(\dot{\theta})$ the yield threshold vanishes, and the units will exhibit rotation already without applied torque, at zero-load rotation speed $\dot{\theta_\textrm{0}} = \eta^{-1}_0(kU - \Gamma_\textrm{f}\mathrm{sgn}(\dot{\theta}))$.
	
	Figure\ \ref{fig_8}D shows that this is consistent with data from experiments where we systematically varied $\Gamma_\textrm{l}$ and measured $\dot{\theta}$ for different voltages $U$. Only very close to zero bias voltage, where the rotor remains at rest until a certain torque load threshold $\Gamma_\textrm{l}^*$ is exceeded, are there deviations. In particular, for the smallest voltages, the angular speed just above $\Gamma_\textrm{l}^*$ is underestimated by the model.
	
	The parameters $R$ and $K_\textrm{e}$ can be determined for each motor from measurements of the motor current performed during the experiments presented in Fig.\ \ref{fig_8}D (see Supplementary Materials and Fig. S4). To determine $K_\textrm{t}$ and $\Gamma_f$, we fit Eq.\ \ref{SteadyMotion} with the data in Fig.\ \ref{fig_8}D. We observe that a single $\Gamma_f$ obtained with such methods allows us to describe well all the behaviors for $\dot{\theta}>0$. However, when $\dot{\theta}$ approaches zero, we observe a threshold that is larger than that estimated by the model. This is consistent with the existence of a static friction torque that is larger than the kinetic one $\Gamma_f$, as commonly observed in dry friction.
	
	\paragraph*{\textbf{Dynamical load response of individual Granulobot units}\\} 
	
	To model the self-oscillating solid regime in Fig.~\ref{fig_3}, we can rewrite Eq. 5 as 
	
	\begin{equation}
		(1+\alpha)\eta_\textrm{0} \dot{\theta} + G\theta = \Gamma_{\textrm{l}} + k(|u| -u^*)\mathrm{sgn}(\dot{\theta}).
		\label{DynamicState}
	\end{equation}
	
	In this form, Eq. 5 shows how in the self-oscillating regime, where $|u_i| -u^* > 0$, the right hand side (RHS) provides a constant force that always pushes the rotor in the direction it is moving. The RHS thus effectively acts as \emph{negative} Coulomb friction, which injects energy rather than taking it out. With this energy input, for large $\alpha$, the angle $\theta$ will approach its maximum excursion while experiencing large viscous, velocity-dependent dissipation, \textit{i.e.} it will exhibit an exponential time dependence similar to charging a capacitor. Near that maximum, where the viscous dissipation will vanish, inherent noise in the system can then trigger a switch of sign in the highly nonlinear sgn function on the RHS of Eq. 9, so that the forcing changes direction and the process reverses, similar to discharging the capacitor. Figure~\ref{fig_8}E shows this behavior (blue) along with the alternating voltage forcing $|u|\mathrm{sgn}(\dot{\theta_i})$ (black). Since the precise moment of switching the direction of forcing  depends on noise, there is chatter in the rise and fall of the forcing, which in turn leads to a small degree of randomness in the successive timing and thus also in the amplitudes of the oscillations (see inset to Fig. ~\ref{fig_8}E). 
	
	As $\alpha$ approaches -1 from above, damping is no longer dominating and dynamic behavior will be reached. We must then take into account the moment of inertia of the rotor $I_\mathrm{\theta}=mr_0^2$, with $m$ the mass of the rotor including the attached magnet, and $r_0$ its radius of revolution. Adding an inertial term to Eq.~5 or Eq.~9 leads to: 
	
	\begin{equation}
		\ddot{\theta} + 2\zeta \omega_0\dot{\theta}+\omega_0^2\theta=\overline{\theta}\omega_0^2 
		\label{eq_dyn}
	\end{equation}
	
	where $\zeta=\frac{(1+\alpha)\eta_0}{2\sqrt{(GI_\mathrm{\theta})}}$ is the damping parameter, $\omega_0=\sqrt{\frac{G}{I_\mathrm{\theta}}}$ is the oscillation frequency of the undamped oscillator, and $\overline{\theta}=\frac{\Gamma_\textrm{l}+k(|u|-u^*)\mathrm{sgn}(\dot{\theta})}{I_\mathrm{\theta}}$ is the characteristic angular displacement under the combined action of the load, the Coulomb friction, and the driving torques. This equation describes a harmonic oscillator with (positive) viscous damping but negative Coulomb friction. It has been used to model driven dissipative systems in different contexts and is known to lead to spontaneous limit cycles \cite{Fulcher2006}.

	For $\zeta>1$, the rotor is in an overdamped regime, resulting in alternating exponential decays (also  described by Eq.9). When the system is turned on, the rotor is initially at rest with $\theta=0$ and $\dot{\theta}=0$. It then starts to oscillate with increasing amplitudes until reaching a steady state. In this regime, Eq.~\ref{eq_dyn} can be solved analytically for a half oscillation, remarking that, as we expect exponential behaviors, we can take as the initial condition $\theta_{n}(t-t_n=0)=\theta_{n-1}(\infty)$ at each half oscillation $n$. Until the angular speed flips sign, the solution of Eq.~\ref{eq_dyn} then takes the form
	
	\begin{equation}
		\theta(t)=\overline{\theta}\left( 1-2e^{\omega_0t(\sqrt{\zeta^2-1}-\zeta)}\right),
		\label{sol_1}
	\end{equation}

	Figure \ref{fig_8}E shows very good agreement between this solution and the experimental data, with no fit parameters. We observe a slight underestimation of the angle when $\dot{\theta}$ approaches $0$, which can be explained by a static friction for vanishing speeds that is larger than $\Gamma_f$ (see Fig \ref{fig_8}D and Fig. S5A for data with larger $\zeta$).
	
	In the intermediate regime, $0<\zeta<1$, inertia starts to be important and we expect the solution giving the half oscillation to overshoot.  As $\zeta$ decreases from $1$ the mechanism that triggers the speed sign flip becomes dominated by the dynamics of the system, hence exhibiting more regular oscillations. The initial conditions for a half oscillation $n$, which corresponds to the first overshoot of the solution at $n-1$, can be determined following the method used in \cite{Fulcher2006} to predict that for large $n$ the oscillations approach a steady-state amplitude $\theta_n \rightarrow \overline{\theta} \left(1+(2e^{\zeta\omega_0\pi/\omega}-1)^{-1} \right)$. Combined with the general form of the solution of Eq.~\ref{eq_dyn} in this regime, this leads for the steady-state to (see details in Supplementary Material):
	
	\begin{equation}
		\theta(t)\simeq\overline{\theta}\left(1-\frac{e^{\frac{-\zeta\omega_0 \pi}{\omega}}}{2\cos\phi}e^{-\zeta\omega_0t}\cos(\omega t+\phi)\right), 
		\label{sol_2}
	\end{equation}
	where $\omega=\omega_0\sqrt{1-\zeta^2}$ and $\tan\phi=-\zeta/\sqrt{1-\zeta^2}$. Figure~\ref{fig_8}F shows good agreement between the analytical prediction and the experimental values for $\zeta=0.47$. The amplitudes are underestimated by the model, but consistent numerically with the observations. 
	
	For $\zeta=0$, \textit{i.e.} $\alpha=-1$ (Fig.~\ref{fig_8}G), there is no longer viscous damping in the system, while energy is constantly added via negative Coulomb friction. If no other balancing mechanism is present, we expect the amplitude $\theta_n$ to grow linearly over time with frequency $\omega=\omega_0=\sqrt{G/I_{\theta}}$, independent of the driving force~\cite{Fulcher2006}. In real system, the total voltage produced by the motor controller cannot exceed a maximum value such that $U\leq U_\textrm{max}$. As a consequence, we observe a saturation of the total voltage $U$ produced by the circuitry (orange curve in Fig.~\ref{fig_8}G), and finite amplitudes at large $n$,  $\theta_n=f(|u|,U_\textrm{max})$ that increase with $|u|$, while $\omega<\omega_0$ decreases with $|u|$ (Fig.~S5B).

	\paragraph*{\textbf{Tensile test experiments}\\}
	
	Tensile test experiments were performed with an Instron 5640 materials tester equipped with a $50$kN load cell. In order to apply strain to Granulobot aggregates, we used two ferromagnetic parallel plates as text fixtures. We then added a third, freely rotating magnet, as in Fig.~S1, to two of the Granulobot units placed on opposite locations within a closed-chain aggregate. To prevent direct contact of these two units' gear surfaces with the Instron plates, we placed slender neodymium magnets between the third rotors and the plates. This firmly attached these two units to the plates while allowing for free rotation of all Granulobot units during testing. All experiments were performed at a constant tensile speed of $1$mm/s.
	
	\paragraph*{\textbf{Effect of Parameter Variability on Granulobot Aggregate States}\\}
	
	In Fig. 3 the parameters $G$ and $|u|$ determine the different states of a Granulobot aggregate for given load.  The values for $G$, which determines the stiffness of the effective torsion spring in each unit, can be programmed into all Granulobot units. In the fluid-like state $G$ is simply set to zero. In the solid-like states, variability in this parameter will affect the effective stiffness of the aggregate but will not affect how the aggregate switches between passive, viscoplastic and self-oscillating solid-like states. On the other hand, the threshold $u^*$ for the voltage bias magnitude $|u|$, which determines the onset of either the active liquid or the active solid, depends on the internal static friction within individual Granulobot units. Variability in this friction, which can come from mechanical components inside the motors and from the frictional motion of all rotors, therefore can have an effect. Going from passive to active flow as $|u|$ is increased,  we can anticipate that some units will initiate motion earlier than others and therefore simply broaden the crossover. The one situation where one could expect a significant effect from variability in the dynamical parameters is the transition to the self-organized self-oscillating solid. However, Fig. 6D shows that the onset of a synchronized state can be observed robustly within a wide range of bias voltages, despite residual hardware variability (we note the slightly differing amplitudes in Fig. 6B,C and Fig. S2E), and without any prior calibration procedures. Furthermore, there is a key benefit from deliberately introducing much larger variability within the self-oscillating regime, for example by choosing different damping parameters among units. In this case we find that aggregates are still self-organizing (Fig. 6E,F,G). While losing synchronization (see Movie S3, Fig. S2B,D), now they exhibit periodic deformations that are global shape changes (Fig 6H) and that form the basis for different self-coordinated locomotion gaits.

	\paragraph*{\textbf{Morphing liquid passing an obstacle}\\}
	
	The onboard monitoring of the rotor angle in each unit makes it possible to develop suitable locomotion strategies by reverse engineering. This proceeds by manually `teaching' the Granulobot aggregate how to overcome a type of obstacle, recording the potentially complex sequence of local angle changes, and from it extracting the essential features required for successful obstacle management.  To do so we follow these three stages:
	\begin{enumerate}
		\item We first put the system in the liquid state, with a low typical damping $\eta^\textrm{0}_{\mathrm{eff}}<0.01$~Ns/rad, which corresponds to $|u_i| = u_0>1.5$~V, such that it behaves as a passive liquid aggregate. We then deform the aggregate  by hand to pass it over an obstacle. Once a continuous sequence of deformations has been  successful in overcoming the obstacle, we reproduce by hand the same sequence without obstacle and record via the Wi-Fi data acquisition platform the associated time-dependent angle changes for each Granulobot unit. 
		
		\item We ``play back'' these angle variations from a central computer to each Granulobot, which they follow via local feedback using the rotor encoder. The voltage signals generated by the feedback loops are recorded. We then use these voltages to infer a general, time-dependent voltage bias $u_i(t)$ appropriate to overcome the type of obstacle just trained on. Figure \ref{fig_7}A shows an aggregate of eight Granulobot units in a closed chain, trained to `flow' over an obstacle.  During training by hand we observe that all eight $u_i(t)$ are oscillating at a similar frequency and amplitude. Furthermore, the signal of a given Granulobot is shifted in phase compared to its neighbor, while Granulobot units placed at opposite positions ($i$ and $i+N/2$) along the chain seem to have the same phase. We generalize this behavior by using a simple periodic function for the bias voltage that shifts each unit's phase by $4\pi/N$ such that   
		$u_\mathrm{i}=(-1)^\mathrm{i}u_\mathrm{0}+A\sin(\omega t+\phi_\mathrm{i})$, with
		$\phi_\mathrm{i+1}=\phi_\mathrm{i}+\frac{4\pi}{N}$ and $A$ the amplitude of the oscillation (shown in Fig. 6A).
		
		\item We send to each Granulobot the address of their neighbor such that they can implement locally the bias $u_\mathrm{i}$ defined in step 2 above, with $u_\mathrm{0}$ and $A$ as user-defined parameters.
	\end{enumerate}
	
	When using such a control strategy, only the pair of parameters ($u_\mathrm{0}$, $A$) must be sent. As the peak voltage $A$ is increased, the aggregate becomes more compliant and follows the contour of the obstacle more closely. This is consistent with a smaller effective viscosity.
	
	\paragraph*{\textbf{Inertial shape-shifter}\\}
	
	The inertial shape-shifter moves by configuring the Granulobot units into a ring-shaped chain and then applies a suitably chosen shape perturbation that propagates around the ring to drive an overall rolling motion, as shown in Fig.\ \ref{fig_7}H and Movie S1. In a closed chain geometry, the sum of angles must satisfy an overall geometric constraint \cite{Chiu2009}, namely
	$\sum_{i=1}^{N}\theta_i(t)=\pi(N-2)$. All angle perturbations $[d\theta_\textrm{1} ... d\theta_i ... d\theta_\textrm{N}]$ of units $[1 ... i ... \textrm{N}]$ must then satisfy $d\theta_i=\sum_{i\neq \textrm{j}}^{N-1}d\theta_\textrm{j}$. To achieve this kinematic control in a  decentralized way, with the same instructions sent to all the Granulobot units, we used a local communication strategy whereby one unit will temporarily become the leader and the others followers. This control strategy can be implemented as a sequence of four sets of instructions that are sent  once  to all units simultaneously at the initial stage:
	\begin{enumerate} 
		\item Learning the initial topological configuration. The sequence of all units' addresses in the ring-like chain is sent, providing units with information about which of their two direct neighbors is the one connected to their active rotor.
		\item Kinematic constraints and initial position of the perturbation. An initial leader address and a number of trailing followers are chosen. This leader and its followers then implement a perturbation as follows: The leader sends in real-time to its actuated neighbor the value of its angle $\theta_i$. This value is propagated through all the followers by neighbor-neighbor communication until the last follower is reached. Every follower applies a rotation that is a linear combination of the leader readout $d\theta_{\textrm{i+1}}=\alpha_{\textrm{i+1}} d\theta_i$, such that the full sequence of coefficient $\{ \alpha_\textrm{1}, ...,\alpha_\textrm{N}\}$ satisfies the angle conservation relationship $d\theta_i=\sum_{i\neq \textrm{j}}^{N-1}d\theta_\textrm{j}$.
		\item Leader unit perturbation. The angle of the leader unit is perturbed according to $d\theta_i(t)=d\theta_{\textrm{0}}\sin(\pi t/T)$, where the parameters $T$ and $d\theta_{\textrm{0}}$ are chosen to control the shape of the perturbation.
		\item Perturbation propagation. When the leader begins the shape perturbation, a timer starts. When this timer reaches $t_{\textrm{com}}$, the leader stops moving and sends a message to its first follower to turn it into the new leader. This triggers a repeat of the perturbation procedure, which then propagates through the chain. 
	\end{enumerate}
	
	The result of this propagating shape perturbation is a shift in the center of gravity of the aggregate that triggers a rolling or crawling motion sustained by the aggregate's inertia.
	
	\section*{Acknowledgments}
	We thank Marc Berthoud for help with software development. This project was funded by NSF grant EFMA-1830939, "EFRI C3 SoRo: Design Principles for Soft Robots Based on Boundary Constrained Granular Swarms". The work utilized shared experimental facilities at the University of Chicago MRSEC, which is funded by the National Science Foundation under award number DMR-2011854.


\begin{thebibliography}{10}
\bibitem{footnote_1}
		Taking as a typical load $\Gamma_{\textrm{l},i} = mgr$, the torque applied to a unit's active rotor of radius $r$ by the weight of a neighboring unit at right angle with respect to gravity, we use Eq. 4 to find $\eta_{\mathrm{eff}} = (1 + \alpha)\eta_\textrm{0} \frac{mgr}{ mgr - k(u^* - |u|)}$.
\bibitem{footnote_2}
		When the $u_i$ across all units have the same sign regardless of $\Gamma_\textrm{f}$, hysteresis is still observed in the regime colored red in Fig.~\ref{fig_3}, but it will be either negative (plasticity) or positive depending on the sign of the load. As a consequence, the transition toward the self-oscillating regime (orange in Fig. 3) can be  shifted slightly. We still observe self-organized oscillating states, but units oscillate sinusoidally regardless of $\alpha$, with a shift of the equilibrium angle $\Delta\theta_0= k(u-u^*)/G$. This is due to alternating within an oscillations between a friction in the direction of the bias, and a friction in the opposite one.
\end{thebibliography}
\end{document}